\documentclass[letterpaper, 10 pt, conference]{ieeeconf}
\usepackage{times}
\usepackage{graphicx}
\usepackage{amsmath,amssymb,amsopn,amstext,amsfonts}
\usepackage{cancel}
\usepackage[space]{cite}
\usepackage{pdfprivacy}
\usepackage{balance}
\usepackage{color}
\usepackage{mathtools}
\usepackage{algpseudocode}
\usepackage{bm}

\usepackage{diagbox}
\usepackage{float}
\usepackage{epstopdf}
\usepackage{pifont}
\usepackage{fixltx2e}
\usepackage{amsmath}
\usepackage{multirow}
\usepackage{url}
\usepackage{verbatim}
\usepackage[linesnumbered,ruled,vlined]{algorithm2e}

\usepackage[linkcolor=black,citecolor=black,urlcolor=black,colorlinks=TRUE]{hyperref}
\usepackage{booktabs}
\usepackage{graphicx}
\usepackage{subcaption}
\usepackage{stfloats}
\usepackage{makecell}
  
\graphicspath{{./fig/}}
\DeclareGraphicsExtensions{.png,.jpg,.eps,.pdf}
\IEEEoverridecommandlockouts
	\IEEEaftertitletext{%
		\begin{center}
			\includegraphics[width=0.98\linewidth]{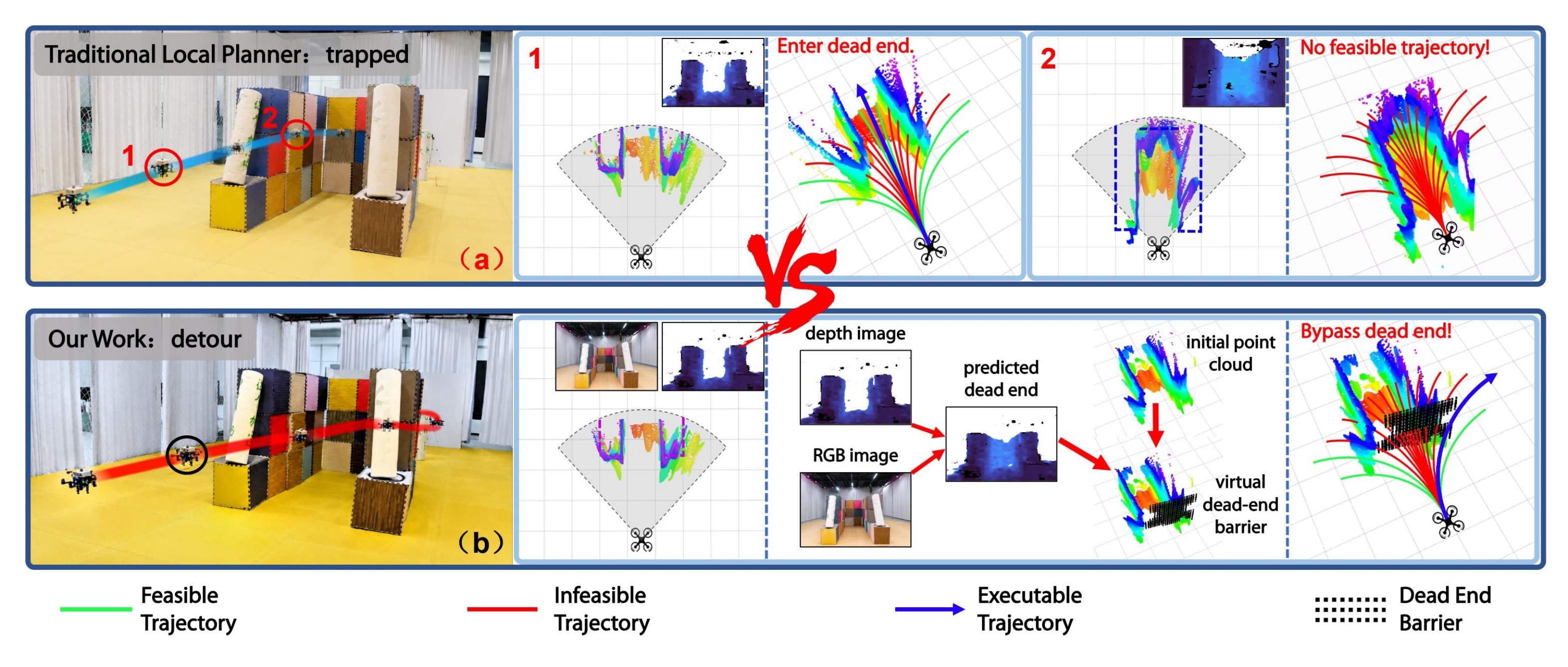}
			\captionsetup{font={small}}
		\captionof{figure}{
			Problem addressed by DPNet: dead-end failures under limited local perception. (a) Depth-only local planning misses the full dead-end structure and fails after entering the trap. (b) DPNet combines RGB semantic cues with depth geometry to predict out-of-FOV dead-end structures before they are fully observed, projects the prediction as virtual barriers, and prunes trajectories leading into the trap, enabling implicit avoidance without dense scene prediction.
		}
		\label{fig:head}
		\vspace{-0.4cm}
	\end{center}
}

		\title{\LARGE \bf DPNet: Efficient Dead-End Prediction and Avoidance for \\ Vision-Based UAV Navigation}

 	\author{Ruibin Zhang$^{*1,2}$, Lun Pan$^{*2,3}$, Zelong Xia$^{2}$, Jialiang Hou$^{1,4,\dagger}$, and Fei Gao$^{1,4,\dagger}$
 		\thanks{This work was supported by the National Key R\&D Program of China under grant No. 2023YFB4706600, the Zhejiang Provincial Science and Technology Plan Project under Grant No. 2024C01170 and the National Natural Science Foundation of China under Grant No. 62322314. (\emph{Corresponding Author: Jialiang Hou; Fei Gao})}
 		\thanks{ $^*$Indicates equal contribution. }
 		\thanks{$^1$Institute of Cyber-Systems and Control, College of Control Science and Engineering, Zhejiang University, Hangzhou 310027, China.}%
 		\thanks{$^2$Huzhou Institute of Zhejiang University, Huzhou 313000, China.}
 		\thanks{$^3$University of Electronic Science and Technology of China, Chengdu 611731, China.}
 		\thanks{$^4$Differential Robotics Technology Company, Hangzhou 311121, China.}
 		\thanks{E-mail: {\tt\small \{ruibin\_zhang, jlhou25\}@zju.edu.cn}}
 	}

\begin{document}
	
	\maketitle
	\thispagestyle{empty}
	\pagestyle{empty}
	\begin{abstract}
	
Vision-based Unmanned Aerial Vehicles (UAVs) often suffer from navigation failures in dead ends due to limited sensing accuracy and range. To address this challenge, this paper proposes a systematic solution for efficient dead-end prediction and avoidance. The proposed method introduces a lightweight neural network to predict the relative distance and bearing of potential dead ends within the current field of view using RGB-D inputs. These predictions prune a predefined, compact trajectory library, enabling the planner to proactively avoid dead ends while maintaining navigational smoothness. Notably, our approach transfers across real-world scenarios without manual annotation or fine-tuning on real-world data. The system achieves high-frequency replanning at 50 Hz onboard. Extensive simulation benchmarks demonstrate superior performance in success rate, flight time, and trajectory length, and real-world experiments further validate its effectiveness in complex scenarios. 
%Our implementation will be released as open source software.

	\end{abstract}
		\vspace{0.4cm} 
	\IEEEpeerreviewmaketitle
	\section{Introduction}
	\label{sec:introduction}
Autonomous navigation in unknown and cluttered environments is a fundamental capability for Unmanned Aerial Vehicles (UAVs), enabling applications ranging from search and rescue\cite{gao2019flying} to industrial inspection\cite{gao2020teach}. To achieve agile and safe flight, UAVs typically use onboard sensors, such as RGB-D cameras or LiDARs, to construct local maps and generate collision-free trajectories. Due to limited onboard computation, gradient-based local planners\cite{zhou2020ego} have become standard, offering high-frequency replanning and smooth trajectory generation from local obstacle information.

However, a critical limitation of these local planners arises from constrained sensing. This is particularly acute for micro UAVs, whose size and payload constraints typically restrict them to visual sensors with limited range and noisy depth estimates. These constraints prevent the planner from observing the full geometry of the environment. Consequently, greedy local planners—which optimize for immediate progress toward a goal—frequently guide the UAV into ``dead ends'' (e.g., U-shaped obstacles or blind alleys). By the time the UAV perceives the blockage, it is often too late to maneuver out, leading to oscillation, collision, or mission failure, as illustrated in Fig.~\ref{fig:head}.

To mitigate the adverse effects of limited onboard sensing, existing methods generally follow two main directions. The first category relies on auxiliary mechanisms outside the local planning loop. This includes triggering recovery maneuvers—such as rotating or backtracking—once the robot is trapped\cite{r3,r4}, or maintaining a global map to search for alternative paths\cite{yang2022far,cao2021tare,dang2019explore}. However, recovery maneuvers are reactive and interrupt flight continuity, while high-resolution global maps are computationally expensive and drift-prone. The second category integrates learning-based prediction into planning\cite{r14,r1,r6,r8}. These methods infer the geometry of unknown or occluded regions, predicting occupancy probabilities or frontiers to guide the robot. While these methods are promising, they often target dense reconstruction or general exploration tasks, which incur high computational latency. Crucially, despite the extensive literature on general obstacle avoidance, proactive dead-end prediction and avoidance remain under-explored. Most planners\cite{zhou2020ego,r3,r4,yang2022far,cao2021tare,dang2019explore} treat dead ends as obstacles requiring reaction rather than structural hazards to anticipate and avoid.

To bridge this gap, we propose DPNet, a system-level solution explicitly designed for efficient dead-end avoidance. Unlike methods that attempt to reconstruct the entire unknown environment\cite{r1,r6,r8}, our approach focuses on the minimal information necessary for obstacle avoidance: identifying the presence, distance, and bearing of dead ends. We use a lightweight encoder-only neural network to predict these hazards from RGB-D inputs and integrate the predictions with a predefined trajectory library. By proactively pruning trajectories that lead into trapped regions, our system ensures safe and continuous navigation without global mapping or reactive recovery maneuvers, achieving high-frequency replanning (50 Hz, including network inference) with onboard edge computing. Moreover, the network is trained exclusively in simulation without requiring manual annotation and demonstrates zero-shot transfer to complex real-world environments. We conduct extensive simulation benchmarks and real-world validation. The results demonstrate that our method outperforms baseline methods\cite{zhou2020ego,r3,r4,hou2025primitiveplanner} by accurately identifying and proactively avoiding dead ends across complex scenarios.

The primary contributions of this paper are as follows:
	\begin{itemize}
			\item [1)] 
		We propose a computationally efficient network for dead-end prediction that achieves zero-shot sim-to-real transfer. It leverages both RGB and depth images—capturing both semantic and structural features—to identify the existence of dead ends as well as their relative distance and bearing.
			\item [2)]
		To the best of our knowledge, we are the first to explicitly address proactive dead-end avoidance for vision-based UAV navigation.
			\item [3)]
		We conduct comprehensive simulation experiments to demonstrate the superiority of our method, and deploy the algorithm on a fully autonomous UAV for validation in complex real-world scenarios. The source code is publicly available to support reproducibility.
%	We plan to open-source our algorithm for community reference.
		\end{itemize}

		\section{Related Work}
	\label{sec:related_works}

 %%%% Motion Planning
		\subsection{Local Planning for UAVs}
	 Local motion planning for UAVs is commonly organized as front-end path planning followed by back-end trajectory optimization \cite{gao2019flying}. The front end generates collision-free geometric paths using search-based methods, sampling-based RRT/RRT* planners \cite{karaman2011sampling}, or motion-primitive formulations such as Hybrid A* \cite{dolgov2010path}. These methods provide completeness guarantees under discretization or sampling assumptions, but become expensive in high-dimensional spaces. The back end then converts the path into a smooth dynamically feasible trajectory, typically using polynomial splines \cite{mellinger2011minimum, richter2016polynomial} or B-splines \cite{zhou2020ego}; collision constraints are often encoded by ESDFs or Safe Flight Corridors \cite{liu2017planning}, while differential flatness further simplifies quadrotor trajectory optimization \cite{wang2022geometrically}.
	 
	 \subsection{Navigation in Unknown Environments}
	 In unknown environments, limited FOV and sensing range lead to partial observability. One response is to maintain global or topological maps: FAR dynamically updates a visibility graph for replanning around dead ends \cite{yang2022far}, TARE builds a sparse global topology for large-scale exploration \cite{cao2021tare}, and Dang et al. use a two-layer graph for subterranean navigation \cite{dang2019explore}. These methods improve global reasoning but add mapping cost and are vulnerable to odometry drift. A second response keeps planning local but adds safety or recovery mechanisms: safe local exploration enforces margins near unknown space \cite{r2}, NanoMap reuses historical depth maps with pose uncertainty \cite{florence2018nanomap}, reactive MPC handles suddenly appearing obstacles \cite{liu2023integrated}, and Faster maintains conservative and aggressive trajectories with the former as backup \cite{r3, r4}. Such mechanisms increase robustness, yet backup or recovery behavior can interrupt smooth flight and increase travel time in cluttered scenes.
	 
 	\begin{figure*}[!t]
		\centering
				\includegraphics[width=0.93\linewidth]{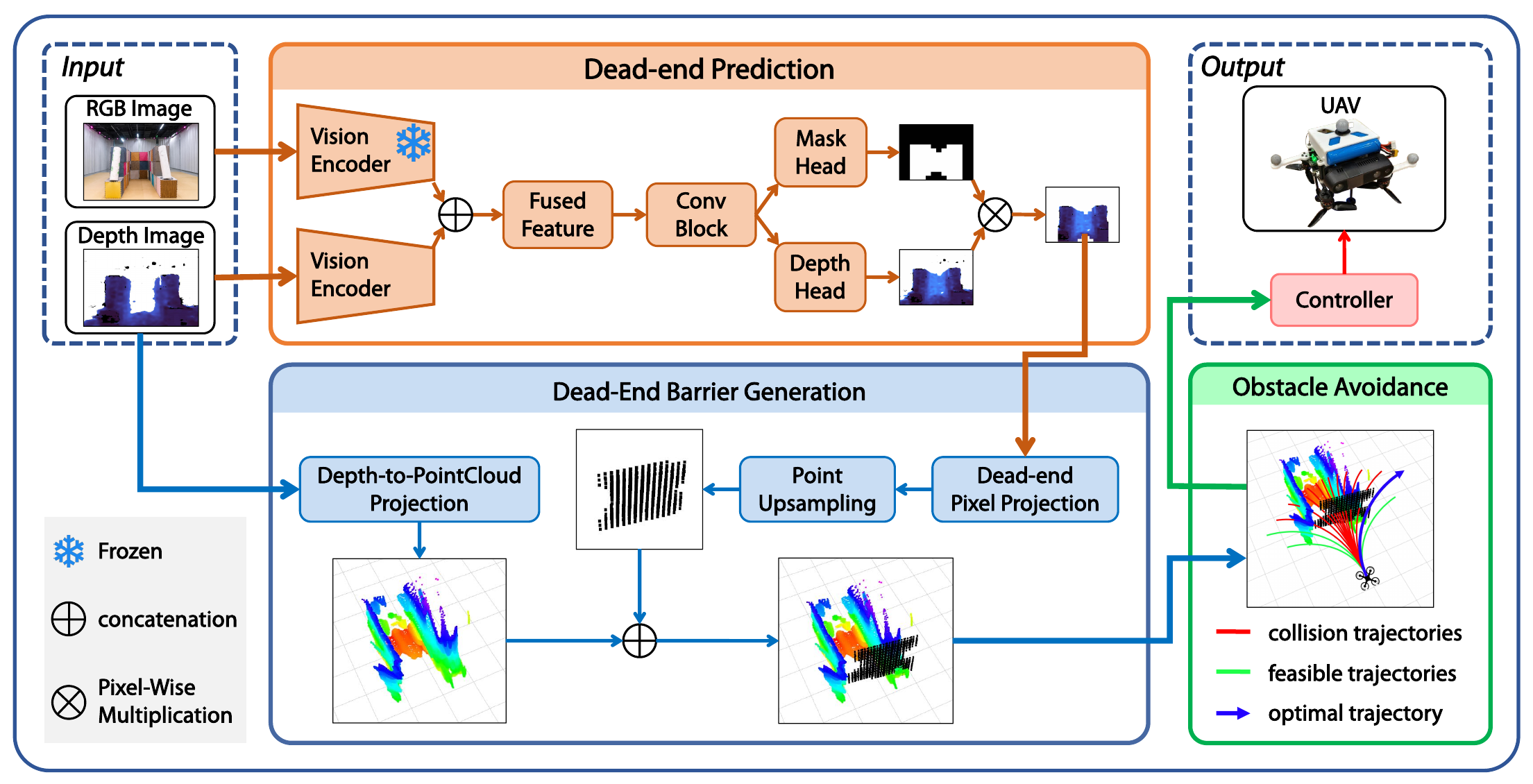}
				\captionsetup{font={small}}
			\caption{System overview of DPNet.}
			\label{fig:system_overview}
	\end{figure*}
% 	\begin{figure*}[!t]
% 	\centering
% 				\includegraphics[width=0.9\linewidth]{fig/network.png}
% 				\captionsetup{font={small}}
% 			\caption{Architecture of the Dead-End Prediction Network.}
% 			\label{fig:network_structure}
% %	\vspace{-0.7cm}
% 	\end{figure*}

\subsection{Predictive Motion Planning}
To mitigate the myopia of reactive local planning, learning-based methods have been introduced to infer information beyond the current sensor horizon. Richter et al.~\cite{r14} predict collision probabilities in unobserved regions, while Nguyen et al.~\cite{nguyen2024uncertainty} estimate spatial uncertainty to guide visually attentive navigation. These probabilistic predictions improve awareness but provide limited geometric constraints for fine-grained obstacle avoidance.

Other methods exploit semantic or geometric prediction for planning. Bartolomei et al.~\cite{bartolomei2020perception} use semantic segmentation to identify hazardous regions, Wang et al.~\cite{r1} predict 3D occupancy in occluded spaces, Song et al.~\cite{song2025p} infer 2D floor plans, and Tao et al.~\cite{tao2023seer} predict information gain for exploration. However, these methods mainly target dense reconstruction or broad exploration objectives, which can be computationally demanding and do not explicitly address proactive dead-end avoidance.

Motivated by these limitations, our work extracts minimal yet actionable dead-end information---presence, distance, and bearing---from RGB-D inputs. Combined with a compact motion primitive library \cite{hou2025primitiveplanner}, this enables efficient trajectory pruning without global mapping or reactive recovery.
		\section{Problem Statement and System Overview}
		\label{sec:system_overview}
			This study considers autonomous UAV navigation in unknown and cluttered environments. The UAV uses a single forward-facing RGB-D camera with limited range and FOV. The objective is to fly from a start to a goal without collision in dead-end environments. We define a dead end as an obstacle structure with concave geometry relative to the UAV's approach direction (e.g., U-shaped, V-shaped, or room corners) and no navigable ``holes'' or topological gaps within the sensor view large enough for safe passage. 
		
		The proposed DPNet framework couples learning-based perception with geometric trajectory planning. As illustrated in Fig.~\ref{fig:system_overview}, RGB and depth images are fed into a network that identifies dead-end regions and outputs a pixel-level mask. To integrate this prediction into the planner, predicted dead-end pixels are projected from 2D image space into 3D. Point upsampling then densifies them into a solid ``virtual obstacle'' barrier that blocks the corresponding path in the planner.
		The obstacle avoidance module then fuses raw depth measurements (physical obstacles) with the upsampled dead-end points into a unified local map. We maintain a fixed-size point cloud to bound computation for real-time performance. Based on this fused point cloud, the planner evaluates a library of motion primitives. Trajectories that collide with either physical obstacles or the predicted dead-end barrier are pruned. The optimal safe trajectory is then sent to the controller for execution.

%  	\begin{figure*}[!t]
% 	\centering
% 				\includegraphics[width=0.9\linewidth]{fig/net_architecture.png}
% 				\captionsetup{font={small}}
% 			\caption{Architecture of the Dead-End Prediction Network.}
% 			\label{fig:network_structure}
% %	\vspace{-0.7cm}
% 	\end{figure*}

	\section{Dead-End Prediction Network}
	\label{sec:network}
	
	\subsection{Network Architecture}
	The proposed network takes an RGB image $\mathbf{I} \in \mathbb{R}^{3 \times H \times W}$ and a depth image $\mathbf{D} \in \mathbb{R}^{H \times W}$, and predicts a downsampled binary mask $\mathbf{\hat{M}} \in [0, 1]^{h \times w}$ and a dead-end depth map $\mathbf{\hat{D}} \in \mathbb{R}^{h \times w}$. The dead-end prediction module in Fig.~\ref{fig:system_overview} shows the network. A dual-branch visual encoder extracts semantic cues from RGB and structural cues from depth, producing aligned features of size $C_{latent} \times h \times w$. The RGB and depth features are concatenated along channels to form $F_{cat} \in \mathbb{R}^{2C_{latent} \times h \times w }$. 	
	
	We then use a dual convolutional block inspired by efficient convolutional designs\cite{r13} to further fuse the multi-modal features. A $1 \times 1$ convolution first compresses the channel dimension from $2C_{latent}$ to a hidden dimension $C_{hidden}$, followed by batch normalization (BN) and ReLU. This step promotes RGB-depth interaction. A subsequent $3 \times 3$ convolution (with padding) further processes local spatial context and reduces the channels to $C_{hidden}/2$.
	Mathematically, the processed feature $F_{proc}$ is obtained as:
	\begin{equation}
	F_{proc} = \text{ReLU}(\text{BN}(\text{Conv}_{3\times3}(\text{ReLU}(\text{BN}(\text{Conv}_{1\times1}(F_{cat})))))).
	\end{equation}
	
	From the processed feature map $F_{proc}$, the network branches into two lightweight parallel heads to generate the task-specific outputs. Both heads adopt a $1 \times 1$ convolution to map the high-dimensional features to single-channel predictions.
	The mask head projects the feature to a dead-end confidence score map. A sigmoid and binarization produce the segmentation mask $\mathbf{\hat{M}}$. The depth head regresses distance values for each pixel, resulting in the depth map $\mathbf{\hat{D}}$. 
	The final output $\mathbf{\hat{O}}$ is generated via a pixel-wise multiplication ($\odot$) of the mask and the depth map:
	\begin{equation}
	\mathbf{\hat{O}} = \mathbf{\hat{M}} \odot \mathbf{\hat{D}}.
	\end{equation}
	This operation ensures that the network outputs valid depth values only for pixels identified as dead ends. If no dead end lies within the current FOV, the all-zero mask, i.e., $\mathbf{\hat{M}} = \mathbf{0}$, leaves the downstream planner operating solely on raw depth measurements without unnecessary constraints.

	\subsection{Dataset Generation}

 	\begin{figure}[t]
	\centering
		\includegraphics[width=0.9\linewidth]{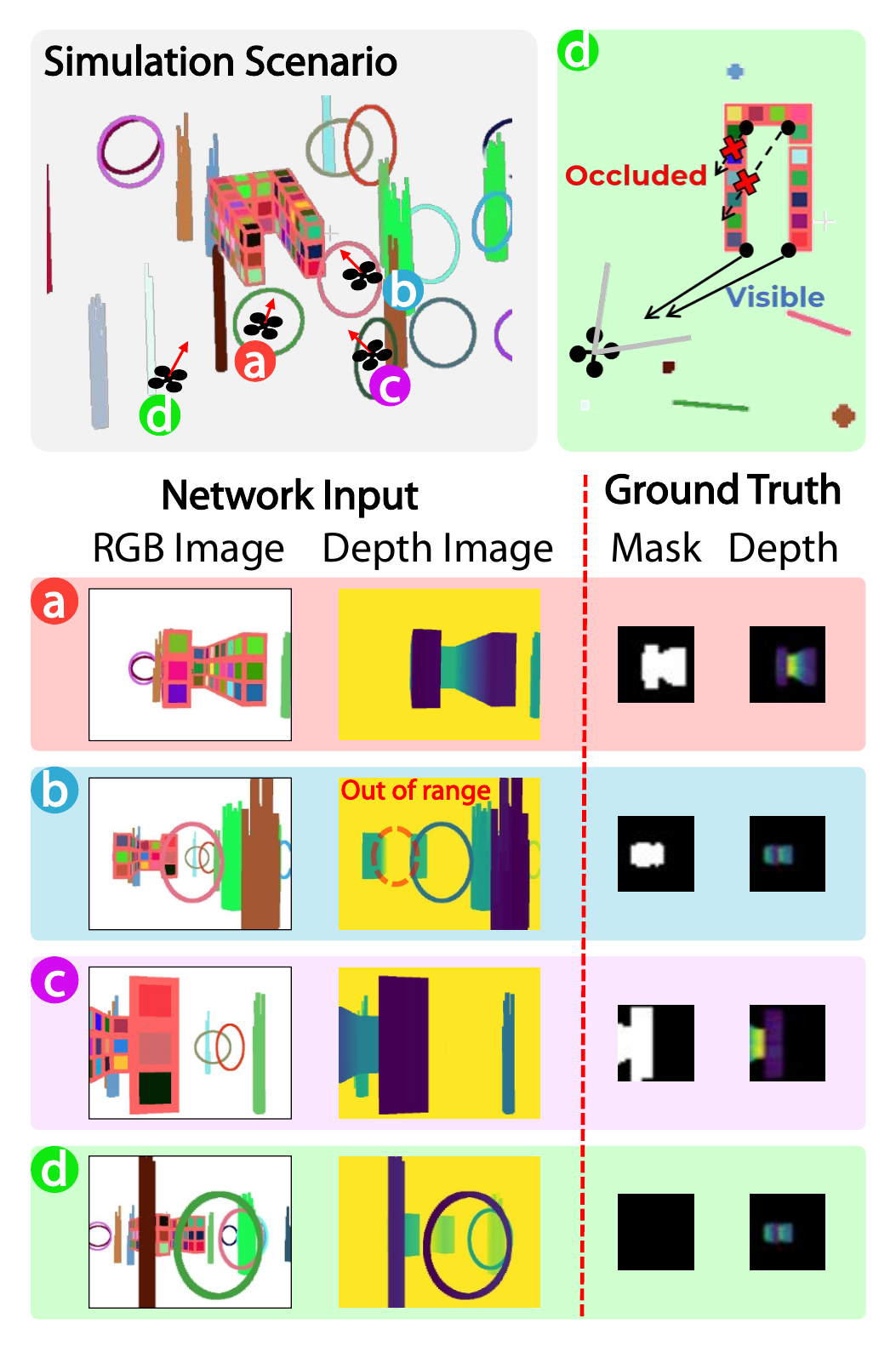}
		\captionsetup{font={small}}
	\caption{Examples of automatic ground-truth labeling. (a) Valid: one inner vertex and all outer vertices are visible, so the dead-end geometry is directly identifiable.
		(b) Valid: the distant dead end is partially clipped in the input depth, but the same geometric visibility criterion still holds.
		(c) Valid: one inner vertex is visible and the out-of-view outer vertex is not occluded, providing sufficient context for shape inference.
		(d) Invalid: no inner vertices are visible, making the structure ambiguous (Mask=0). 
	}
	\label{fig:dataset_generation_illustration}
	%	\vspace{-0.7cm}
	\end{figure}

	Training data acquisition in real-world environments is labor-intensive and prone to labeling errors. To address this, we construct a simulation scenario to generate a diverse auto-labeled dataset, as illustrated in Fig.~\ref{fig:dataset_generation_illustration}. We simulate cluttered environments with random generic obstacles (e.g., cylinders, boxes, and rings) and dead-end structures. In this study, we consider two common topological types of dead ends: U-shape and V-shape. We cap dead ends with a virtual ceiling to enforce non-traversability.
	To ensure the dataset covers a wide range of flight conditions, we randomly sample 3D camera poses ($p \in \mathbb{R}^3, q \in SO(3)$) within the free space of the simulation scenario.
		For each sampled pose, we project the global scene point cloud onto the image plane to generate RGB images with obstacle-specific colors and depth images from the Euclidean distance to the camera; values exceeding $D_{max}$ are clipped to simulate limited sensor range.

%	 RGB images are obtained by rendering the scene with distinct colors for different objects. Depth Images are obtained by calculating the Euclidean distance from the camera center to the projected points. Crucially, to simulate the limited sensing range of real-world depth cameras, we apply a clipping operation to the input depth map. Any depth value exceeding a threshold $D_{max}$ is set to $D_{max}$.
	
	The ground truth consists of two components: the dead-end mask $\mathbf{M}$ and depth $\mathbf{D}$. All ground-truth images are resized from the input resolution $H \times W$ to the network output resolution $h \times w$.
	
	$\mathbf{D}$ represents the geometric structure of the dead end. We generate it by projecting the complete dead-end point clouds onto the image plane. Unlike input depth, it is not clipped. This design forces the network to infer the complete structure and true depth of the dead end, even when parts are beyond the sensor range or occluded in the input data.
	
	$\mathbf{M}$ determines whether a dead end is visible in the current view. We propose an efficient geometry-based auto-labeling algorithm to replace manual annotation. We simplify a dead end as a non-closed polygon in top view, with inner vertices $\mathcal{V}_{in}$ and outer vertices $\mathcal{V}_{out}$ (the endpoints of the non-closed edge). 
		For each vertex $v$, we project it onto the image plane and classify it as \textbf{Out-of-View} if it lies outside the image boundaries, \textbf{Occluded} if it is within the FOV but deeper than the corresponding scene depth value, and \textbf{Visible} if it is within the FOV and not blocked.
	A dead end is considered \textbf{identifiable} if it satisfies either of the following conditions:
	(a) All inner vertices $\mathcal{V}_{in}$ are visible. (b) At least one inner vertex is visible, and all outer vertices $\mathcal{V}_{out}$ are not occluded (i.e., visible or out-of-view, but not blocked by other obstacles). If identifiable, the binary mask $\mathbf{M}$ is generated by projecting the dead-end point cloud (pixel values set to 1); otherwise, $\mathbf{M}$ is an all-zero matrix.

\begin{algorithm}[t]
	\caption{Automatic Ground Truth Generation for Dead Ends}
	\label{alg:auto_labeling}
	\SetKwInOut{Input}{Input}
	\SetKwInOut{Output}{Output}
	\SetKwFunction{FCheck}{CheckVisibility}
	\SetKwProg{Fn}{Function}{:}{}
	
	\Input{
		Point cloud of the dead end $P_{cloud}$; 
		Inner/Outer vertices $\mathcal{V}_{in}, \mathcal{V}_{out}$; 
		Camera pose $T_{wc}$; 
		Scene depth map $D_{scene}$; 
		Camera intrinsics $K$
	}
	\Output{Dead-end mask $\mathbf{M}$, dead-end depth $\mathbf{D}$}
	
	\BlankLine
	% Define the Visibility Check Function
	\Fn{\FCheck{$v, T_{wc}, K, D_{scene}$}}{
		$p_{img}, d_{v} \leftarrow \text{Project}(v, T_{wc}, K)$\;
		\If{$p_{img}$ is outside image boundaries}{
			\Return \textbf{OUT\_OF\_VIEW}\;
		}
		$d_{map} \leftarrow D_{scene}[p_{img}]$\;
		\If{$d_{v} > d_{map} + \epsilon$}{
			\Return \textbf{OCCLUDED}\;
		}
		\Return \textbf{VISIBLE}\;
	}
	
	\BlankLine
	% Main Process
	\Fn{\text{GenerateGT}}{
		$S_{in} \leftarrow [ \FCheck(v) \mid v \in \mathcal{V}_{in} ]$\;
		$S_{out} \leftarrow [ \FCheck(v) \mid v \in \mathcal{V}_{out} ]$\;
		
		\BlankLine
		\tcp{Check visibility conditions}
		$C_1 \leftarrow \forall s \in S_{in}, s = \textbf{VISIBLE}$\;
		$C_2 \leftarrow (\exists s \in S_{in}, s = \textbf{VISIBLE}) \land (\forall s \in S_{out}, s \neq \textbf{OCCLUDED})$\;
		
		\BlankLine
		\eIf{$C_1 \lor C_2$}{
%			\tcp{Dead end is identifiable}
			$\mathbf{D} \leftarrow \text{ProjectCloud}(P_{cloud}, T_{wc}, K)$\;
			$\mathbf{M} \leftarrow \text{Binarize}(\mathbf{D})$\;
		}{
%			\tcp{Dead end is invisible or ambiguous}
			% $\mathbf{D} \leftarrow \mathbf{0}_{h \times w}$\;
			$\mathbf{D} \leftarrow \text{ProjectCloud}(P_{cloud}, T_{wc}, K)$\;
			$\mathbf{M} \leftarrow \mathbf{0}_{h \times w}$\;
		}
		\Return $\mathbf{M}, \mathbf{D}$\;
	}
\end{algorithm}

	\subsection{Training and Employment}
	\label{sec:training_and_employment}
		We employ a multi-task objective that jointly optimizes dead-end depth prediction and mask classification. Depth regression is supervised by the $L1$ loss only over valid dead-end regions, while mask prediction uses the binary cross-entropy (BCE) loss. The weighted loss $\mathcal{L}$ is:
 
\begin{equation}
	\begin{split}
		\mathcal{L} = \;& \lambda_d \lVert \mathbf{M} \odot (\mathbf{\hat{D}} - \mathbf{D}) \rVert_{1} \\
		& - \lambda_m \left[ \mathbf{M} \log(\mathbf{\hat{M}}) + (1-\mathbf{M}) \log(1-\mathbf{\hat{M}}) \right], 
	\end{split}
	\label{eq:loss}
\end{equation}
		where $\lambda_{d}$ and $\lambda_{m}$ are weights for the two loss terms. 
		
		The RGB and depth encoders are initialized with pre-trained DINOv3 weights\cite{simeoni2025dinov3}, leveraging feature representations learned from large-scale datasets. Specifically, we adopt an asymmetric fine-tuning strategy:
		\begin{itemize}
		\item
		RGB Branch: The encoder weights are frozen during training. Since the RGB encoder is pre-trained on diverse real-world images, it serves as a robust feature extractor, bridges the sim-to-real domain gap, and prevents overfitting to simulation-specific textures.
		\item
		Depth Branch: The single-channel depth input is replicated three times to match the three-channel input requirement of the pre-trained encoder. Unlike the RGB branch, the depth encoder is fine-tuned to adapt to depth-map geometry.
		\end{itemize}	
		
		The 50,000 synchronized RGB-D image-label pairs are randomly split into training, testing, and validation sets at 7:2:1. The model is trained for 100 epochs on a single NVIDIA RTX 4090 GPU using Adam with an initial learning rate of $5 \times 10^{-5}$ and a batch size of 128.
		All input images are resized to $H \times W = 224 \times 224$. Since the vision encoder uses a patch size of 16, the resulting feature map and output prediction have a spatial resolution of $h \times w = 14 \times 14$. Although coarse-grained compared with pixel-level dense prediction, this resolution provides sufficient spatial context for effective dead-end avoidance, as demonstrated in the experiments.

		After obtaining the raw network prediction $\mathbf{\hat{O}} = \mathbf{\hat{M}} \odot \mathbf{\hat{D}}$, we post-process it for downstream planning.
		First, a statistical outlier removal filter eliminates isolated noise pixels.
		The remaining valid pixels are then projected into 3D space. To reliably block the path and prevent the planner from traversing the dead end, we construct a virtual ``barrier'' at the dead-end entrance by assigning the minimum predicted depth value to all valid pixels. This flattens the geometry into a planar envelope parallel to the image plane. The 3D coordinates $\mathbf{p}_i \in \mathbb{R}^3$ for each valid pixel $(u_i, v_i)$ are:

%	\begin{equation} \mathbf{p}_i = d_{min} \cdot K^{-1} \begin{bmatrix} u_i \ v_i \ 1 \end{bmatrix},  \text{with } d_{min} = \min \mathbf{\hat{O}}(u_i, v_i),
%	\label{eq:projection} 
%	\end{equation}

\begin{equation}
	\begin{split}
		\mathbf{p}_i &= d_{min} \cdot K^{-1} \begin{bmatrix} u_i & v_i & 1 \end{bmatrix}^T, \\
		\text{with }  d_{min} &= \min_{(u_i, v_i) \in \mathcal{P}_{valid}} \mathbf{\hat{O}}(u_i, v_i),
	\end{split}
	\label{eq:projection}
\end{equation}
		where $K$ denotes the camera intrinsic matrix, and $\mathcal{P}_{valid}$ is the set of pixel coordinates where the predicted mask is active ($\mathbf{\hat{M}}(u_i, v_i) = 1$). Finally, we adopt Moving Least Squares (MLS)\cite{alexa2003computing} to upsample the point cloud by a factor of 10 and obtain $P_{deadend}$. This yields several hundred points, forming a dense obstacle boundary for the planner.
	
	\section{Obstacle Avoidance with Dead-End Awareness}
	\label{sec:obstacle_avoidance}
			This section presents the proposed dead-end-aware trajectory selection strategy. The module enables the UAV to avoid immediate obstacles and proactively reject paths leading to local minima while remaining computationally efficient.

 	\begin{figure}[t]
	\centering
				\includegraphics[width=0.88\linewidth]{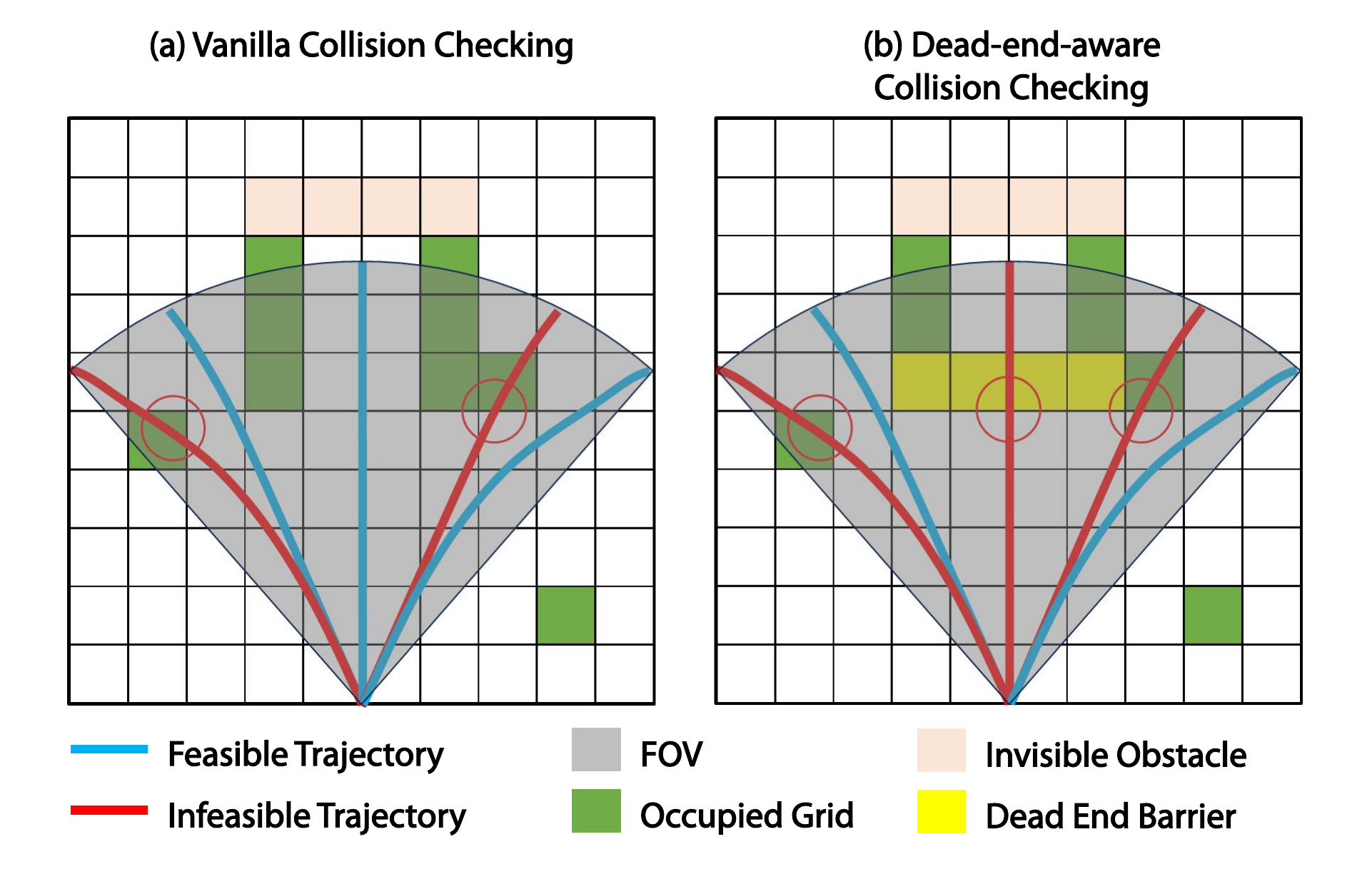}
				\captionsetup{font={small}}
			\caption{Dead-end-aware collision checking.}
	\label{fig:collision_checking}
	%	\vspace{-0.7cm}
	\end{figure}
	
 	\begin{figure}[t]
	\centering
				\includegraphics[width=0.85\linewidth]{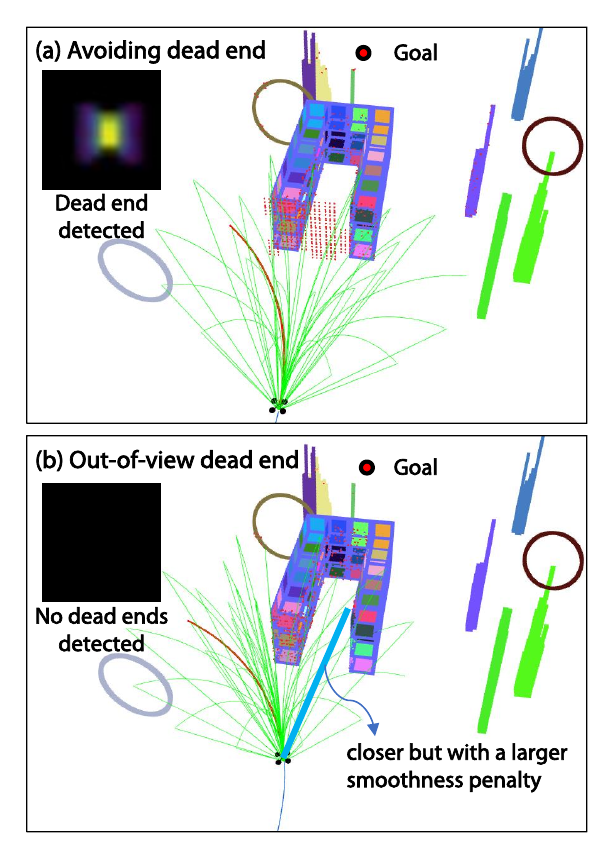}
				\captionsetup{font={small}}
			\caption{Dead-end-aware trajectory selection. The two subfigures are consecutive snapshots. (a) The UAV bypasses the dead end. (b) After the dead end leaves the field of view, the smoothness penalty prevents the planner from selecting a trajectory that re-enters the dead end.}
	\label{fig:trajectory_selection}
	%	\vspace{-0.7cm}
	\end{figure}

	\begin{figure*}[t]			
	\centering
	\includegraphics[width=0.92\linewidth]{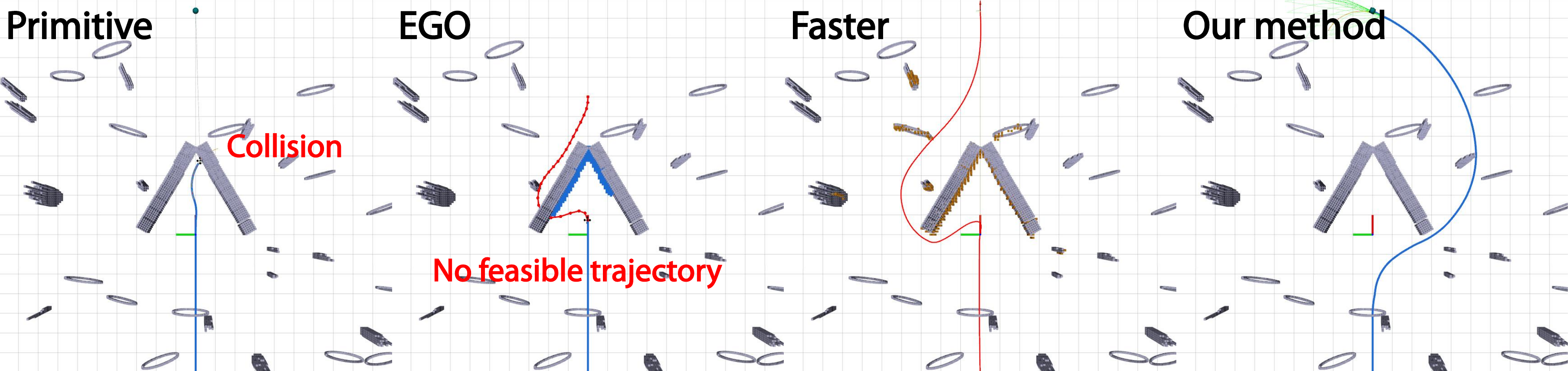}
	\captionsetup{font={small}}
	\caption{Planner behavior comparison in a representative simulation scenario.}
	\label{fig:sim_comparison}
	\end{figure*}		

	\subsection{Collision Checking}
	\label{sec:collision_checking}
		We use a precomputed time-optimal trajectory library generated offline\cite{hou2025primitiveplanner}. For safety, each trajectory is collision-checked against the current environmental perception.
		Following the efficient indexing mechanism proposed by Primitive-Planner \cite{hou2025primitiveplanner}, we maintain the local coverage space of each trajectory as fixed-size virtual voxel grids. During the online phase, the collision check verifies whether the environment point cloud occupies any voxel grid associated with a candidate trajectory. Thus, checking time depends only on the fixed number $N_{fixed}$ of downsampled points. To integrate dead-end avoidance, we fuse the raw depth point cloud $P_{raw}$ with the virtual dead-end barrier points $P_{deadend}$ (generated in Section~\ref{sec:training_and_employment}) to form a unified obstacle set $P_{total}$.
	Crucially, $P_{total}$ is then downsampled to the same fixed size $N_{fixed}$. Since checking still uses $N_{fixed}$ points, dead-end constraints impose \textbf{zero} additional computational overhead over the vanilla collision check. As shown in Fig.~\ref{fig:collision_checking}, trajectories leading into the dead end are pruned because they collide with the virtual barrier.
	
	\subsection{Trajectory Selection}
	\label{sec:trajectory_selection}
		After pruning invalid paths, the optimal trajectory is selected from the remaining trajectory set $\mathcal{T}_{valid}$ by minimizing a composite cost function. The baseline planner \cite{hou2025primitiveplanner} evaluates trajectories based on the goal cost ($c_{goal}$), which penalizes distance to the global goal, and the boundary cost ($c_{bound}$), which penalizes trajectories exceeding the map boundary:
	\begin{equation}
		c_{base} = w_g c_{goal} + w_b c_{bound}, 
		\label{eq:primitive_planner_base}
	\end{equation}
		where $w_{g}$ and $w_{b}$ are the corresponding weights. We refer readers to Primitive-Planner~\cite{hou2025primitiveplanner} for details on $c_{goal}$ and $c_{bound}$. 
		However, in dead-end scenarios, using $c_{base}$ alone is insufficient. When the UAV initiates a detour to avoid a dead end, the structure may temporarily leave the limited FOV. Once the virtual barrier disappears from the instantaneous observation, the greedy nature of $c_{goal}$ can induce the planner to select a trajectory that sharply turns back toward the global goal (and consequently, back into the dead end), as illustrated in Fig.~\ref{fig:trajectory_selection}. To address this ``re-entry'' issue, we introduce a smoothness consistency cost ($c_{smooth}$) to penalize sudden angular changes. We discourage trajectories whose heading direction $\psi_{curr}$ deviates significantly from the previous trajectory heading $\psi_{prev}$:
	\begin{equation}
		\begin{split}
			& \Delta \psi = \lVert \psi_{curr} - \psi_{prev} \rVert _{1}, \\
			& c_{smooth} = 
			\begin{cases} 
				\Delta \psi, & \text{if } \Delta \psi > \psi_{thresh} \\
				0, & \text{otherwise}
			\end{cases}
		\end{split}
		\label{eq:smooth_cost}
	\end{equation}
		where $\psi_{thresh}$ is a relaxation threshold allowing necessary maneuvers. The final objective function is $c_{total} = c_{base} + w_s c_{smooth}$. Finally, the controller executes the selected trajectory for safe navigation.

	\section{Results}
	\label{sec:Results}

	\subsection{Implementation Details} 
\label{sec:Lightweight Verification}
 
	 Experiments are conducted on a custom-built quadrotor equipped with an Intel RealSense D435 RGB-D camera, which provides time-aligned color and depth images at VGA resolution. Invalid depth pixels caused by stereo holes near edges, reflective surfaces, or distant regions are filled using Navier--Stokes-based inpainting before point-cloud projection. The UAV pose is provided by a motion capture system to isolate algorithmic performance from state-estimation errors, and the selected polynomial trajectory is sent to a PX4 flight controller as desired position, velocity, acceleration, and yaw commands.
	 
	 Onboard computation runs on an NVIDIA Jetson Orin NX. DPNet inference is accelerated using TensorRT with FP16 precision and takes 9.2\,ms on average; dead-end point-cloud post-processing and trajectory planning take 2.4\,ms and 5.1\,ms, respectively. The complete pipeline achieves a stable control frequency of \textbf{50\,Hz}. Simulation experiments use camera intrinsics and image resolutions identical to those of the real platform to ensure consistent sensing conditions.

	\subsection{Benchmark and Experiment}
\label{sec:Benchmark_Comparisons}
 
	 To evaluate the effectiveness of our method under limited local perception, we first conduct simulation benchmarks against three representative open-source local planners and then validate real-world performance. Faster~\cite{r3, r4} maintains aggressive and conservative trajectories with the latter as a backup, EGO-Planner~\cite{zhou2020ego} is a gradient-based local planner without explicit safety fallback, and Primitive-Planner~\cite{hou2025primitiveplanner} generates trajectories from a pre-computed motion primitive library without online optimization.
	 	 
	 \textbf{Simulation Benchmark.}
	 Each method is evaluated over 100 trials under two difficulty levels in unknown RGB-D scenes with random pillars, rings, U-/V-shaped dead ends, and II-shaped through-passages as negative cases. The \textit{easy} setting uses 15 pillars, 15 rings, one dead end, and velocity/acceleration limits of 2\,m/s and 8\,m/s$^2$; the \textit{hard} setting increases the obstacle count to 25 and the limits to 3\,m/s and 12\,m/s$^2$. A trial succeeds if the robot reaches within 0.3\,m of the goal without collision; collision or timeout from infeasibility is counted as failure. Table~\ref{tab:benchmark} reports the success rate, flight time, and trajectory length.
 
\begin{table}[t]
   \centering
  \footnotesize
   \caption{Benchmark comparison under limited local perception.}
   \label{tab:benchmark}
   \setlength{\tabcolsep}{5pt}
   \begin{tabular}{lcccc}
     \hline
     \thead{Diff.} & \thead{Method} & \thead{Succ.\\(\%)} & \thead{Time\\(s)} & \thead{Length\\(m)} \\
     \hline
     \multirow{4}{*}{Easy} & EGO-Planner & 80.0 & 14.54 & \textbf{23.00} \\
                           & Primitive-Planner & 86.0 & 19.91 & 29.90 \\
                           & Faster & 97.0 & 18.47 & 25.83 \\
                           & \textbf{Ours} & \textbf{100.0} & \textbf{13.77} & 24.35 \\
     \hline
     \multirow{4}{*}{Hard} & EGO-Planner & 76.0 & 11.65 & \textbf{22.13} \\
                           & Primitive-Planner & 47.0 & \textbf{9.06} & 22.34 \\
                           & Faster & 96.0 & 13.27 & 26.31 \\
                           & \textbf{Ours} & \textbf{99.0} & 9.93 & 24.92 \\
     \hline
   \end{tabular}
 \end{table}
 
	\begin{figure*}[!t]			
	\centering
	\includegraphics[width=0.97\linewidth]{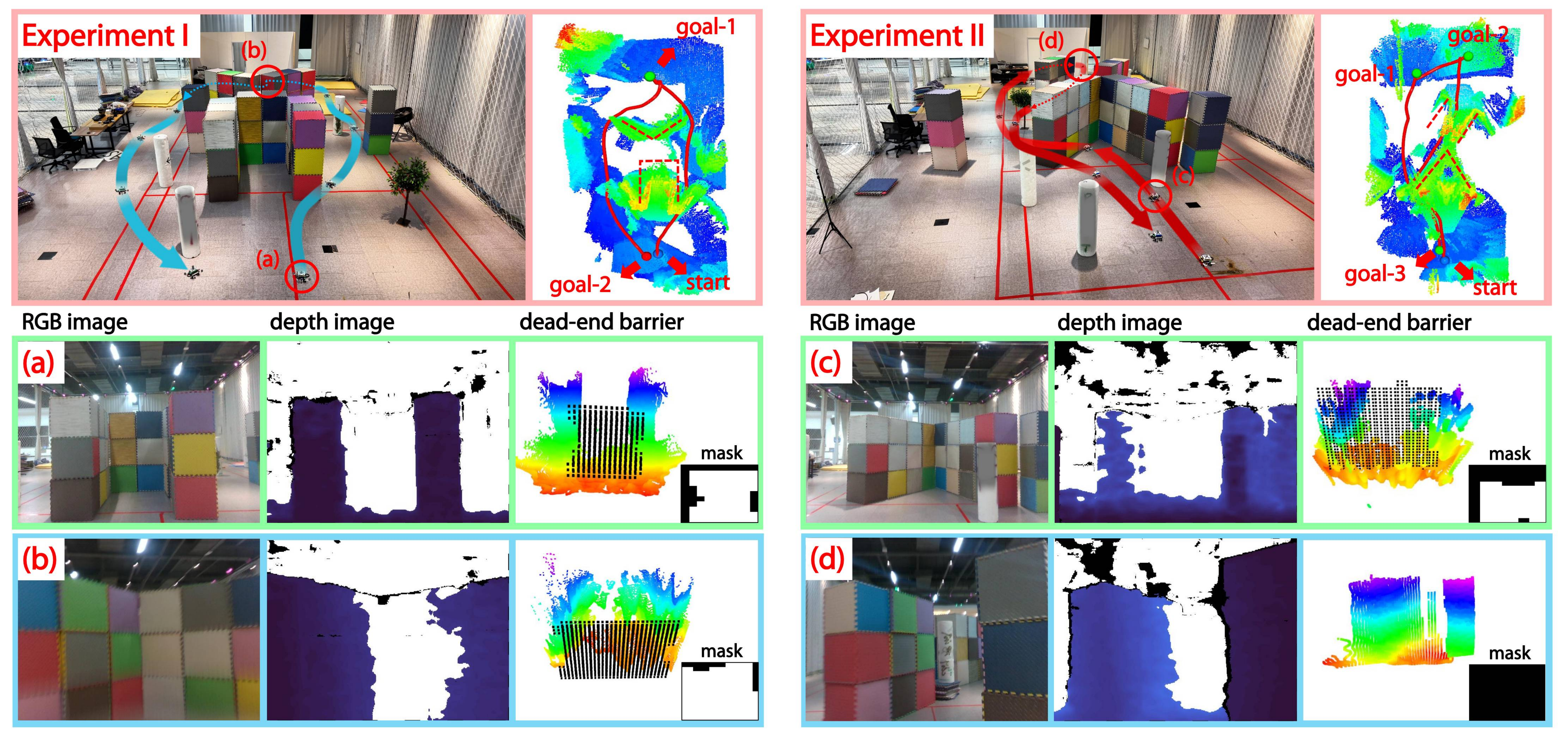}
	\captionsetup{font={small}}
	\caption{The real-world experiments. (a)--(c) Key observations of true dead ends. (d) The opened control passage used for comparison.}
	\label{fig:real_exp_result}
	\end{figure*}	

 Table~\ref{tab:benchmark} shows that our method achieves the highest success rate in both settings, reaching 100.0\% in the easy case and 99.0\% in the hard case. Compared with Faster, it also shortens the average flight time by 25.4\% and 25.2\% and yields shorter trajectories. This indicates that proactive dead-end pruning avoids the recovery overhead caused by entering locally unfavorable regions. Although EGO-Planner and Primitive-Planner sometimes report shorter time or length on successful trials, these statistics exclude their failures; for example, Primitive-Planner succeeds in only 47.0\% of hard trials. Therefore, the combined metrics indicate a better robustness-efficiency tradeoff for our method.
 
 Fig.~\ref{fig:sim_comparison} illustrates a representative case. Primitive-Planner and EGO-Planner follow the locally attractive entrance and collide or fail to find a feasible trajectory. Faster avoids failure through recovery but produces backtracking and detouring. In contrast, our method predicts the dead end before entry, inserts a virtual barrier, and directly selects an outer bypass route.
	
	\subsection{Real-world UAV Navigation}
	\label{sec:Real-world UAV Navigation}

	 To demonstrate and verify DPNet in real-world environments, we conduct two indoor navigation experiments in a motion-capture arena using the same onboard system without real-world fine-tuning, as shown in Fig.~\ref{fig:real_exp_result}. Both experiments are cluttered with dead-end obstacles and several regular obstacles. In Experiment II, an additional control obstacle is placed, where the terminal side of a U-shaped structure is opened to form a traversable passage. The UAV is required to navigate a series of sequential goals. These goals are specially arranged so that the UAV has to pass near the dead-end entrances and detour around them, rather than avoiding the challenging regions by chance. Thus, the experiments test both proactive dead-end avoidance and discrimination of passable look-alikes. Depth inputs are clipped at 4\,m, and the maximum velocity and acceleration are set to 1\,m/s and 5\,m/s$^2$.

	 As can be seen from Fig.~\ref{fig:real_exp_result}(a)--(c), the depth images only partially capture the dead-end structures at the decision points, and the terminal closures cannot be reliably perceived from local geometry. Nevertheless, DPNet predicts masks for these dead-end regions. The corresponding virtual obstacle point clouds are then inserted near the entrances, forcing the planner to reject trajectories that enter the trapping regions. As a result, the UAV safely bypasses the true dead ends and reaches the assigned goals. In contrast, the control passage in Fig.~\ref{fig:real_exp_result}(d) is not identified as a dead end; no virtual barrier is generated, and the UAV passes through it normally. The point-cloud top views in Fig.~\ref{fig:real_exp_result} clearly visualize the virtual obstacle point clouds and the resulting UAV trajectories, including detours around true dead ends and direct traversal through the control passage. The above results reveal that DPNet maintains stable prediction under noisy, incomplete RGB-D sensing and enables proactive avoidance while preserving traversable passages.

\balance
		\section{Conclusion} 
		\label{sec:conclusion}

	 In this work, we focus on proactive dead-end avoidance for vision-based UAV navigation. We present DPNet, a task-oriented perception-planning framework. Instead of reconstructing the full unknown environment, DPNet predicts compact dead-end masks and depths from RGB-D observations. Based on these predictions, virtual barriers are generated and fused with local depth measurements, allowing the motion-primitive planner to prune trajectories leading to trapping structures. Extensive simulation benchmarks validate the proposed method's robustness and efficiency compared with representative planners. Real-world experiments further demonstrate that DPNet can transfer from simulation to physical scenes and distinguish true dead ends from traversable look-alikes. The above results show that compact, task-oriented prediction can effectively complement local planning under limited onboard perception. In future work, we plan to investigate UAV navigation in more challenging environments, such as intricate dead-end layouts, scenarios with multiple concurrent dead ends, and maze-like structures.

		\newlength{\bibitemsep}\setlength{\bibitemsep}{0.0\baselineskip}
	\newlength{\bibparskip}\setlength{\bibparskip}{0pt}
	\let\oldthebibliography\thebibliography
	\renewcommand\thebibliography[1]{%
		\oldthebibliography{#1}%
		\setlength{\parskip}{\bibitemsep}%
		\setlength{\itemsep}{\bibparskip}%
	}
	\bibliography{tmech2025_zrb}

\end{document}